\documentclass[conference]{IEEEtran}  

\IEEEoverridecommandlockouts                              

\usepackage{cite}  
\usepackage{amsmath} 
\usepackage{amssymb}  

\usepackage{graphicx}
\usepackage{capt-of} 
\usepackage{overpic} 
\usepackage[export]{adjustbox} 
\usepackage{minibox} 
\usepackage{textcomp} 
\usepackage{tabularx}
\usepackage{multirow}
\usepackage{booktabs}
\usepackage{arydshln}
\usepackage{adjustbox} 
\usepackage{pgfplots} 
\usepackage{siunitx}
\usepackage[implicit=false]{hyperref}

\usepackage{algorithm}
\usepackage{algpseudocode}
\algblock{Input}{EndInput}
\algnotext{EndInput}
\algblock{Output}{EndOutput}
\algnotext{EndOutput}
\newcommand{\Desc}[2]{\State \makebox[2em][l]{#1}#2}

\usepackage[caption=false, font=scriptsize]{subfig} 

\newcommand\copyrighttext{\footnotesize \textcopyright~2023 IEEE. Personal use of this material is permitted. Permission from IEEE must be obtained for all other uses, in any current or future media, including reprinting/republishing this material for advertising or promotional purposes, creating new collective works, for resale or redistribution to servers or lists, or reuse of any copyrighted component of this work in other works.
}%

\newcommand\copyrightnotice{%
	\begin{tikzpicture}[remember picture,overlay]
	\node[anchor=south,xshift=0pt,yshift=14pt] at (current page.south) {\fbox{\parbox{\dimexpr\textwidth-\fboxsep-\fboxrule\relax}{\copyrighttext}}};
	\end{tikzpicture}%
}

\begin{document}
\title{RESET: Revisiting Trajectory Sets for Conditional Behavior Prediction
\thanks{The research leading to these results is funded by the BMWK within the project "KI Delta Learning" (F\"orderkennzeichen 19A19013A).}}

\author{\IEEEauthorblockN{Julian Schmidt\IEEEauthorrefmark{1}\IEEEauthorrefmark{2}, Pascal Huissel\IEEEauthorrefmark{1}, Julian Wiederer\IEEEauthorrefmark{1}\IEEEauthorrefmark{3}, Julian Jordan\IEEEauthorrefmark{1}, Vasileios Belagiannis\IEEEauthorrefmark{3} and Klaus Dietmayer\IEEEauthorrefmark{2}}
	\IEEEauthorblockA{\IEEEauthorrefmark{1}Mercedes-Benz AG, Research \& Development, Stuttgart, Germany\\
		Email: julian.sj.schmidt@mercedes-benz.com}
	\IEEEauthorblockA{\IEEEauthorrefmark{2}Ulm University, Institute of Measurement, Control and Microtechnology, Ulm, Germany}
	\IEEEauthorblockA{\IEEEauthorrefmark{3}Friedrich-Alexander-Universität Erlangen-Nürnberg, Erlangen, Germany}}
\maketitle

\begin{abstract}
It is desirable to predict the behavior of traffic participants conditioned on different planned trajectories of the autonomous vehicle.
This allows the downstream planner to estimate the impact of its decisions.
Recent approaches for conditional behavior prediction rely on a regression decoder, meaning that coordinates or polynomial coefficients are regressed.
In this work we revisit set-based trajectory prediction, where the probability of each trajectory in a predefined trajectory set is determined by a classification model, and first-time employ it to the task of conditional behavior prediction.
We propose RESET, which combines a new metric-driven algorithm for trajectory set generation with a graph-based encoder.
For unconditional prediction, RESET achieves comparable performance to a regression-based approach.
Due to the nature of set-based approaches, it has the advantageous property of being able to predict a flexible number of trajectories without influencing runtime or complexity.
For conditional prediction, RESET achieves reasonable results with late fusion of the planned trajectory, which was not observed for regression-based approaches before.
This means that RESET is computationally lightweight to combine with a planner that proposes multiple future plans of the autonomous vehicle, as large parts of the forward pass can be reused.
\end{abstract}


\section{Introduction}

\begin{figure}[t]
	\centering
	\includegraphics[clip, trim=0.0cm 0cm 0.0cm 0.0cm, width=\linewidth]{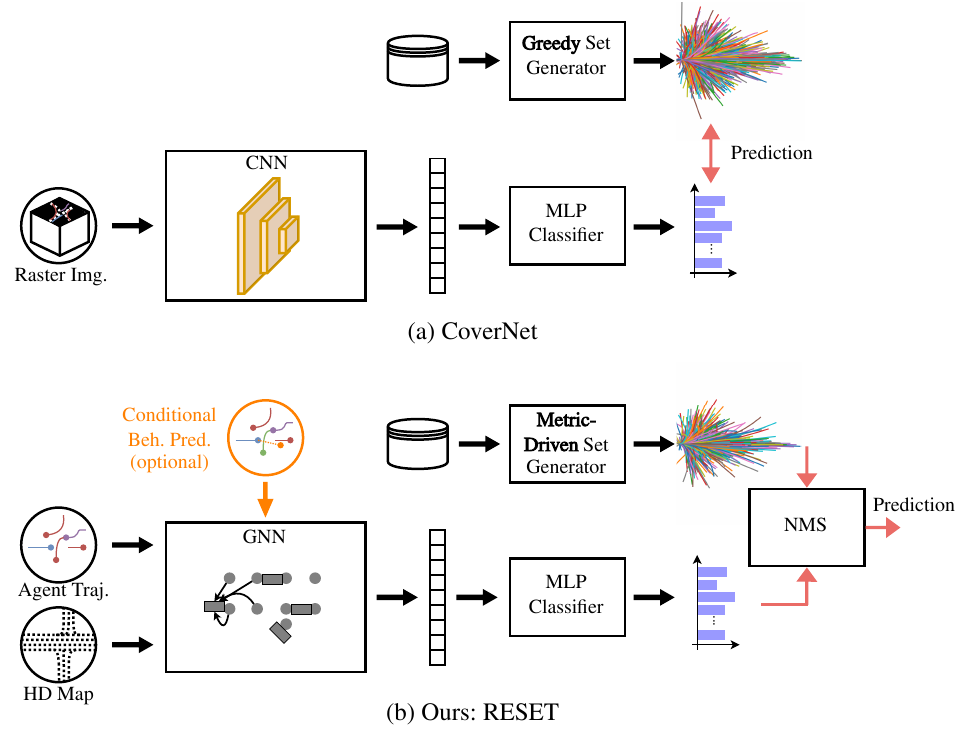}
	\caption{Comparison of (a) CoverNet and (b) our method RESET: We use a new metric-driven algorithm for set generation, a graph-based input representation and a Non Maximum Suppression post-processing strategy. The model is also applicable to the task of conditional behavior prediction.}
	\label{fig:teaser}
\end{figure}

Safe trajectory planning in complex traffic scenarios requires the autonomous vehicle to predict the future motion of surrounding traffic participants.
Lately, machine learning-based prediction models, e.g.,~\cite{Liang2020, Gao2020, Gu2021, Zeng2021}, have shown promising results, as they are able to efficiently leverage information provided by perception and fusion stacks in combination with information from a High Definition (HD) map.
For highly interactive scenarios, it is desired to additionally condition the predictions on the planned future trajectory of the autonomous vehicle.
In this way, a downstream planner can estimate the impact of its planned trajectories on surrounding agents, as predictions resulting from multiple planned trajectories can be compared to each other.
\copyrightnotice 

In order to make conditional behavior prediction computationally lightweight during planning, it is advantageous to inject the future trajectory of the autonomous vehicle in later stages of the model.
Thereby, during the forward pass, the calculations up to the fusion point can be reused for multiple possible planned futures.
Prior publications~\cite{Huang2022_ARXIV, Tolstaya2021} that compare early fusion (fusion during encoding) and late fusion (fusion prior to decoding) with regression decoders share one observation:
Early fusion always outperforms late fusion by a margin, with late fusion even leading to worse predictions than without conditioning.

In this paper we do not follow recent approaches, e.g.,~\cite{Liang2020, Lee2017, Zeng2021}, that typically predict coordinates or polynomial coefficients during a regression step.
Instead, we revisit set-based trajectory prediction and employ it to the task of conditional behavior prediction.
Set-based trajectory prediction, originally proposed in CoverNet~\cite{PhanMinh2020}, reformulates the task of trajectory prediction as a classification problem.
Rather than regressing trajectories, the probability of each trajectory in a trajectory set is predicted.
Without retraining and influencing the runtime or complexity, this allows, up to the size of the used set, a flexible amount of trajectories with a high probability being used as predictions.
On benchmarks, CoverNet fails to achieve competitive results with respect to other recent approaches.

Therefore, this paper is two-fold:
Firstly, we propose a set-based trajectory prediction method that achieves similar performance to regression-based approaches.
We name this method \textit{RESET} (REvisiting SET-based trajectory prediction).
RESET consists of a new metric-driven algorithm for trajectory set generation, a graph-based encoder and a Non Maximum Suppression (NMS) post-processing strategy.
Fig.~\ref{fig:teaser} highlights the differences to CoverNet.
Secondly, we show that our method is employable to the task of conditional behavior prediction and evaluate different variants of how to inject the conditioning information into the model.
While late fusion fails for regression decoders in prior work, our set-based method achieves reasonable results with late fusion.

In summary, our main contributions are:
\begin{itemize}
    \item We revisit set-based trajectory prediction and propose a method that achieves similar performance to a regression-based model.
    \item We compare different injection points for conditioning the set-based prediction on the planned future trajectory of the autonomous vehicle.
    \item We extensively evaluate our set-based method for unconditional and conditional prediction on the large-scale Argoverse Motion Forecasting Dataset~$2$~\cite{Wilson2021} and show that it obtains reasonable results with late fusion.	
\end{itemize}

\section{Related Work}
This section discusses the related work in trajectory prediction, with a special focus on goal- and set-based approaches.
Additionally, we refer to prior work related to conditional behavior prediction.

\subsection{Trajectory Prediction}
Trajectory prediction is a constantly evolving research topic with a myriad of different approaches.
One group of approaches directly regresses trajectories via a regression decoder~\cite{Liang2020, Gao2020, Zhou2022, Schmidt2022}.
Adding an additional classification decoder allows determining the probability per predicted trajectory~\cite{Liang2020, Zhou2022}.
Distribution-based approaches, on the other hand, are used to regress trajectories from a latent distribution, most commonly realized with Conditional Variational Auto-encoders (CVAEs)~\cite{Lee2017, Rhinehart2018, Tang2019, Casas2020} or Generative Adversarial Networks (GANs)~\cite{Gupta2018}.
In contrast to direct regression-based approaches, when using CVAEs with a continuous latent space, it is also possible to predict agents with an arbitrary amount of modes, without having to predefine this amount prior to training~\cite{Lee2017}.
This is also a strength of goal- and set-based approaches, which are discussed in-depth below.

\textbf{Goal-based approaches}
first sample goal candidates and then score these over-sampled candidates in order to obtain endpoint predictions~\cite{Gu2021, Zeng2021, Wang2022}.
Goal candidates can either be sparsely predefined anchors~\cite{Zeng2021, Wang2022} or densely sampled~\cite{Gu2021}.
To obtain trajectory predictions, an additional regression step is required.
This regression step is commonly performed in a learning-based manner and uses the top-scored goal candidates.

\textbf{Set-based approaches}
do not share the similarity of all above mentioned approaches, which is the dependency on a regression decoder.
Instead, they initially generate a trajectory set containing a variety of possible trajectories and then use a classifier to determine the most likely trajectories of the generated set.
PRIME~\cite{Song2021} relies on a model-based planner to generate a trajectory set for the agent to be predicted.
The generated set is based on the current state of the agent and the HD map.
A subsequent learning-based classification network is then used to determine which trajectories the agent will most likely drive.
This results in a two-stage approach.

CoverNet~\cite{PhanMinh2020} uses a rasterized representation of agents and map and proposes two variants of trajectory sets: The dynamic and the fixed set.
The dynamic set is generated by the rollout of a kinematic model (e.g., kinematic bicycle model) using diverse control actions.
This kinematic model is initialized with the current state of the agent to be predicted.
Similar to PRIME, this results in a two-stage approach, meaning that an initial model-based set generation is required prior to the learning-based classification.
The fixed set, on the other hand, is the same for each agent.
It is already generated from the training split before the training process and inducts prior knowledge about the dataset into the model.
For this set generation, a greedy bagging algorithm is used to determine the minimal set of trajectories in a dataset covering all other trajectories.
Whether a trajectory covers another trajectory is determined by checking whether the maximum point-wise $\ell^2$ distance is below an acceptable error tolerance.
Due to the fixed set being generated from the training dataset prior to the training process, this results in a one-stage approach.

Our approach is also one-stage, making use of fixed trajectory sets.
In contrast to the rasterization-based encoding of agents and map and the greedy bagging algorithm for trajectory set generation of CoverNet, we combine a new metric-driven set generation algorithm, a graph-based encoder and an NMS post-processing strategy.

\subsection{Conditional Behavior Prediction}
Injecting additional information about the future of one or multiple agents into the prediction model is called conditioning.
Most commonly, predictions are conditioned on the planned future trajectory of the autonomous vehicle.
Conditional behavior prediction has been examined in the context of direct regression-based models~\cite{Tolstaya2021, Khandelwal2020_ARXIV, Ngiam2022} and distribution-based models~\cite{Rhinehart2019, Salzmann2020}.
However, it has never been investigated, whether goal-based and set-based prediction approaches are also capable of using the additional information that is injected during the conditioning process.

In this work we cover this research gap for set-based approaches and investigate different ways of employing our set-based method to the task of conditional behavior prediction.

\section{Method}
\begin{figure*}[tpb]
	\centering
	\includegraphics[width=\textwidth]{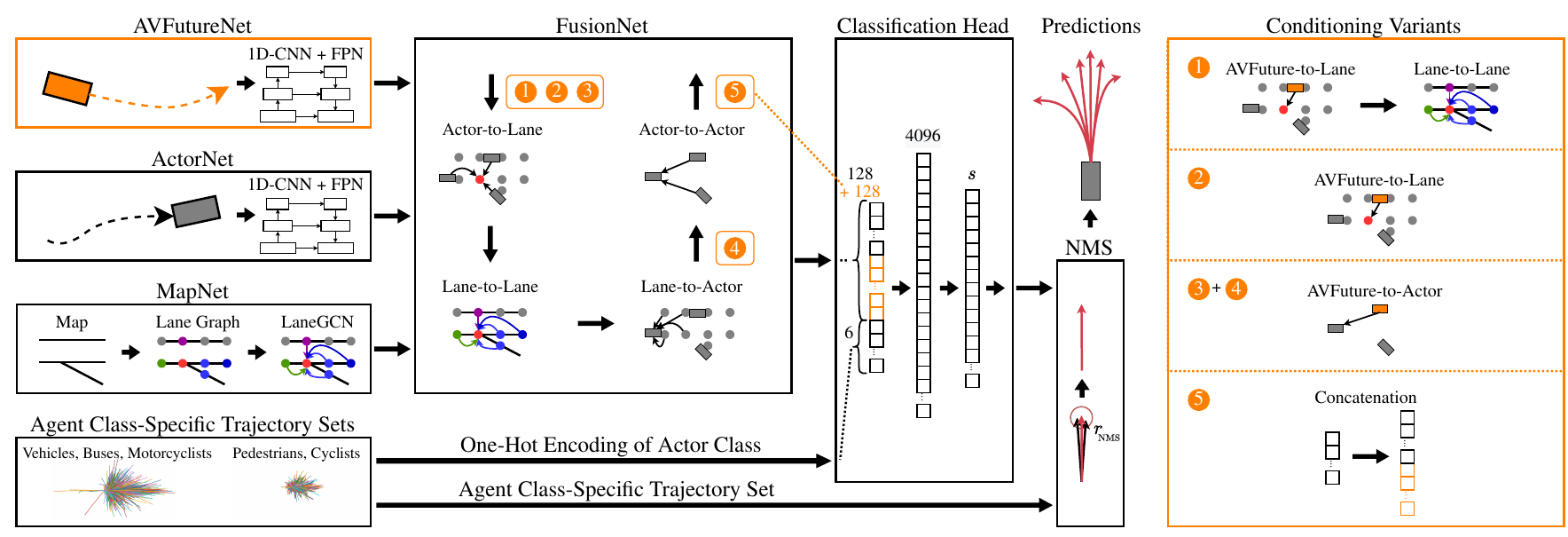}
    \vspace{-0.7cm}
	\caption{Overview of RESET: Information of past agent motion (ActorNet) and the HD map (MapNet) are fused in four fusion stages (FusionNet). Subsequently, the classification layer determines a probability for each trajectory in the trajectory set. Combining these probabilities with the trajectory set leads to trajectory predictions, which are post-processed with a Non Maximum Suppression. Orange blocks indicate different variants for conditioning the predictions on the planned future trajectory of the autonomous vehicle (encoded by the AVFutureNet). Five variants, ranging from early to late fusion, are examined.}
	\label{fig:method_full}
\end{figure*}

This section describes our method RESET, consisting of the trajectory set generation, the encoder-decoder classification model and an NMS post-processing strategy.
An overview is given in Fig.~\ref{fig:method_full}.
Note that this overview already includes different variants for conditioning the predictions on the future trajectory of the autonomous vehicle, colored in orange.
Details are given in Section~\ref{sec:method_conditional}.

In general, trajectory prediction is defined as the task of predicting the future trajectory $\mathbf{T}_\mathrm{gt} = \{ \boldsymbol{\tau}^t \}^{T_f}_{t=1}$ of an agent.
$T_f$ defines the prediction horizon and $\boldsymbol{\tau}^t = (x^t, y^t)$ the position of an agent at timestep $t$ in a 2D Bird's Eye View (BEV) coordinate system.

\subsection{Trajectory Set Generation}
We propose to use a metric-driven algorithm for trajectory set generation and to use different trajectory sets for different agent classes.
During the training and inference, these trajectory sets are used as the classification target by the decoder.
Prior to trajectory set generation, all considered trajectories are transformed into a common local coordinate system.
Only future timesteps $t=1, 2, \dots ,T_F$ are considered.

\subsubsection{Algorithm}
\begin{figure}[!t]
\vspace{-0.3cm}
\begin{algorithm}[H]
	\small
	\caption{Metric-driven trajectory set generation}
	\label{alg:set_generation}
	\begin{algorithmic}[1]
	  	\Input
		\Desc{$\mathcal{D}$}{Dataset containing trajectories $\mathbf{T}_i$, $|\mathcal{D}| = k$}
		\Desc{$s$}{Final set size}
		\EndInput
		\Output
		\Desc{$\mathcal{S}$}{Trajectory set, $|\mathcal{S}| = s}$
		\EndOutput
		\Procedure{get\_traj\_set}{$\mathcal{D}$, $s$}
		\State $\mathcal{S} \leftarrow \{\}$
		\State $\mathbf{M}_{\mathrm{set}} \leftarrow [\inf, \inf, ..., \inf] \in \mathcal{R}^k$  \label{alg:set_generation:m_set_init}
		\State $\mathbf{M}_\textrm{full} \in \mathcal{R}^{k\times k}$, where $\mathbf{M}_\textrm{full}[i, j] \leftarrow \Call{get\_ade}{\mathbf{T}_i, \mathbf{T}_j}$ \label{alg:set_generation:m_full_init} 
		\While{$|\mathcal{S}| < s$} \label{alg:set_generation:main_loop}
			\State Initialize empty metric memory set $\mathcal{M}_\mathrm{memory} \leftarrow \{\}$
			\State $i \leftarrow 0$
			\While{$i < k$}
				\State $\mathbf{M}_{i, \mathrm{combined}} \leftarrow [\mathbf{M}_\textrm{full}[i], \mathbf{M}_{\mathrm{set}}] \in \mathcal{R}^{k\times 2}$ \label{alg:set_generation:concat_temporary}
				\State $\mathbf{M}_i \leftarrow \min\limits_{\mathrm{axis}=1}{\mathbf{M}_{i, \mathrm{combined}}} \in \mathcal{R}^k$  \label{alg:set_generation:M_i}
				\State $m_i \leftarrow \mathrm{mean}\,(\mathbf{M}_i) \in \mathcal{R}$ \label{alg:set_generation:m_i}
				\State $\mathcal{M}_\mathrm{memory} \leftarrow \mathcal{M}_\mathrm{memory} \cup \{(i, m_i, \mathbf{M}_i)\}$
				\State $i \leftarrow i+1$
			\EndWhile
		\State $(i, m_i, \mathbf{M}_i) \leftarrow \min\limits_{m_i}{\mathcal{M}_\mathrm{memory}}$ \label{alg:set_generation:min_traj_extraction}
		\State $\mathbf{M}_{\mathrm{set}} \leftarrow \mathbf{M}_i$
		\State $\mathcal{S} \leftarrow \mathcal{S} \cup \{ \mathbf{T}_i \}$ \label{alg:set_generation:set_appending}
		\EndWhile
		\State \Return $\mathcal{S}$
		\EndProcedure
		\\
		\Procedure{get\_ade}{$\mathbf{T}_i$, $\mathbf{T}_j$}
			\State \Return $\frac{1}{T_f} \sum_{t=1}^{T_f} ||\boldsymbol{\tau}^{t}_i - \boldsymbol{\tau}^{t}_j||_2$
		\EndProcedure
	\end{algorithmic}
\end{algorithm}
\vspace{-0.8cm}
\end{figure}
Our proposed algorithm, shown in Algorithm~\ref{alg:set_generation}, is iterative and driven by a selected prediction metric.
In our case, we use the minimum Average Displacement Error (minADE).
Other metrics, such as minFDE or MR are also applicable.
More details on these metrics are given in Section~\ref{subsec:metrics}.

The input to the algorithm are the dataset $\mathcal{D}$ and the desired final set size $s$.
After the defined number of iterations $s$, the algorithm outputs the trajectory set $\mathcal{S}$.

During an initialization phase, the ADE value of each trajectory pair in the dataset $\mathcal{D}$ is calculated and stored in $\mathbf{M}_\mathrm{full}$ (Line~\ref{alg:set_generation:m_full_init}).
$\mathbf{M}_\mathrm{set}$ is initialized with arbitrary high values (Line~\ref{alg:set_generation:m_set_init}).
During the iterative set generation process, $\mathbf{M}_\mathrm{set}$ will store the minADE for each trajectory in the dataset $\mathcal{D}$ with regard to the current trajectory set $\mathcal{S}$.

In each iteration (defined by the loop in Line~\ref{alg:set_generation:main_loop}), one trajectory $\mathbf{T}_i$ is added to the trajectory set $\mathcal{S}$.
The trajectory $\mathbf{T}_i$ is selected in an optimal way with regard to the metric, in our case minADE.
This is ensured by the following procedure:
Each trajectory $\mathbf{T}_i$ in the dataset $\mathcal{D}$ is temporarily appended to the trajectory set (Line~\ref{alg:set_generation:concat_temporary}).
$\mathbf{M}_i$ is calculated to store the minADE of each trajectory in the dataset $\mathcal{D}$ with regard to the current temporary trajectory set (Line~\ref{alg:set_generation:M_i}).
Averaging $\mathbf{M}_i$ results in one scalar $m_i$ that corresponds to the average minADE that is optimally achievable with the current temporary trajectory set on the dataset $\mathcal{D}$ (Line~\ref{alg:set_generation:m_i}).
In each iteration, the one trajectory $\mathbf{T}_i$ that yields the lowest optimally achievable minADE $m_i$ (Line~\ref{alg:set_generation:min_traj_extraction} to Line~\ref{alg:set_generation:set_appending}) is picked and added to the trajectory set $\mathcal{S}$.

\subsubsection{Agent Class-Specific Trajectory Sets}
We also propose to use agent class-specific trajectory sets.
The motivation is simple: The motion of an agent is highly dependent on its class. For instance, vehicles typically move faster and according to a different kinematic model than pedestrians.
Having predefined trajectory sets for specific agent classes therefore inducts prior knowledge into the prediction model.
Algorithm~\ref{alg:set_generation} is still used for trajectory set generation, but the dataset $\mathcal{D}$ is limited to trajectories of agents with selected classes only.

\subsection{Encoder} \label{subsec:encoder}
Following the principle of the Graph Neural Network (GNN)-based LaneGCN~\cite{Liang2020}, we use a graph-based encoding for lane centerlines and past agent trajectories:
MapNet uses layers of LaneGCN in order to extract lane node features.
In the ActorNet, a 1D-Convolutional Neural Network (CNN) and Feature Pyramid Network (FPN) are used to extract features from the past trajectories of agents.
These trajectories are represented as a series of displacement vectors.
We extend this representation by concatenating a one-hot encoding of the corresponding agent's class.
Information of map and agents are combined in the FusionNet, a stack of four interaction blocks, namely Actor-to-Lane, Lane-to-Lane, Lane-to-Actor and Actor-to-Actor.
Output of the encoder is one feature vector per agent.
This feature vector is aware of the motion and position of surrounding agents (social context) and the lane geometries (static context).

It is important to emphasize that the encoder is not the focus of this work.
We intentionally only use the series of displacement vectors, the agent classes and the lane centerlines as input information to our model.
This allows for a fair comparison to the baselines introduced in Section~\ref{subsec:baselines}.
As demonstrated in a prior publication~\cite{Zhang2022_ARXIV}, our encoder is extendable to making use of more detailed lane information and a full tracked state, including velocity and orientation.
The publication also shows that making use of such information leads to an improvement in prediction performance.

\subsection{Decoder}
The classification head is a Multilayer Perceptron (MLP) with a $\mathrm{softmax}$ output layer.
The output layer size corresponds to the number of trajectories in the trajectory set.
Input to the MLP are the features resulting from the encoder.
Output is a probability for each trajectory in the trajectory set.
During inference, up to the size of the used trajectory set, a flexible and arbitrary amount of $k$ top-scored trajectories can be used as predictions.

\subsection{Non Maximum Suppression}
To increase the diversity in the predictions, we make use of an NMS strategy.
During inference, the trajectory with the highest probability is picked first and all trajectories with endpoints that lie within a radius $r_\mathrm{NMS}$ of the first trajectory's endpoint are discarded.
This procedure is repeated until the desired number of modes $k$ are obtained.

\section{Employing the Model to Conditional Behavior Prediction} \label{sec:method_conditional}
RESET is the first set-based approach that is able to perform conditional behavior prediction, where additional information about the future trajectory of the autonomous vehicle is injected.
Fig.~\ref{fig:method_full} highlights the blocks used for this purpose in orange.
Similar to ActorNet, the future trajectory of the autonomous vehicle is encoded with a 1D-CNN and a FPN (AVFutureNet).
Input to AVFutureNet is the future trajectory of the autonomous vehicle, transformed to the local coordinate system used by ActorNet and MapNet.
We examine five different ways and stages (Variant~$1$ to $5$) of fusing this information with the features of predicted agents.
Variant $1$ to $3$ are based on early fusion, because information about the future trajectory of the autonomous vehicle is injected prior to map encoding.
Variant $4$ and $5$ are based on late fusion, as a maximum of one Actor-to-Actor block and the MLP decoder are applied after the fusion.

\textbf{Variant $\mathbf{1}$}
is a lane-based fusion.
Information about the future trajectory of the autonomous vehicle is first fused into the lane nodes, using an AVFuture-to-Lane block.
AVFuture-to-Lane uses the same attention-based fusion that is used in the Actor-to-Lane block in later stages of the fusion cycle.
Subsequently, the Lane-to-Lane mechanism distributes this information across the lane graph.

\textbf{Variant $\mathbf{2}$}
uses an AVFuture-to-Lane interaction block that is similar to the Actor-to-Lane block used subsequently.

\textbf{Variant $\mathbf{3}$ and $\mathbf{4}$}
both use an AVFuture-to-Actor interaction block that is similar to the Actor-to-Actor block.
Variant~$3$ does this in an early fusion manner, meaning that the encoded information of AVFutureNet is initially fused to the agent-wise features resulting from ActorNet.
Variant~$4$ does this in a late fusion manner, meaning that the encoded information is fused subsequent to the Actor-to-Lane, Lane-to-Lane and Lane-to-Actor blocks.

\textbf{Variant $\mathbf{5}$}
is concatenation-based, meaning that the features resulting from AVFutureNet are concatenated to the features of each agent resulting from FusionNet.
This results in a larger input layer of the classification head.

\section{Experiments}
This section describes the evaluation of our model, including the different variants for conditional behavior prediction, on the publicly available Argoverse Motion Forecasting Dataset~$2$~\cite{Wilson2021}.

\subsection{Dataset}
The Argoverse~$2$ dataset contains \SI{199908}{} sequences for training, \SI{24988}{} for validation and \SI{24984}{} for testing.
Each sequence consists of an HD map and a \SI{11}{s} recording of detected and tracked agents in the surrounding of an autonomous vehicle, which is sampled with \SI{10}{Hz}.
Given $\SI{5}{s}$ of a recording, the goal is to predict the motion of the remaining $\SI{6}{s}$ of one selected agent, namely the focal agent.

\subsection{Metrics} \label{subsec:metrics}
Evaluation is done using standard metrics for single- ($k=1$) and multi-modal ($k=6$) predictions.
The minimum Average Displacement Error (minADE) is the average Euclidean distance between the predicted trajectory and the ground-truth trajectory.
Similarly, the minimum Final Displacement Error (minFDE) is the Euclidean distance between the predicted endpoint and the ground-truth endpoint.
For multi-modal \mbox{minADE} and minFDE evaluation ($k>1$), only the prediction with the smallest Euclidean endpoint distance to the ground-truth is considered.
minADE and minFDE are averaged over all sequences in the dataset.
Miss Rate (MR) is the ratio of sequences where none of the $k$ predicted endpoints lies within a radius of $\SI{2}{m}$ of the ground-truth endpoint.

Turn Rate Infeasibility (TRI)~\cite{Cui2020} is used to evaluate the percentage of trajectories that are kinematically infeasible in terms of their turn radius.
Including the last observed position, predicted trajectories with a radius smaller than an agent class-specific threshold are labeled as infeasible and vice versa\footnote{Used turn radius thresholds: vehicle: \SI{1.8}{m}, bus: \SI{3.0}{m}, motorcyclist: \SI{0.8}{m}, cyclist: \SI{0.6}{m} and pedestrian: \SI{0.0}{m}.}.

\subsection{Implementation Details}
The model is trained with cross-entropy as a loss function.
The classification target is given by the trajectory in the trajectory set that is closest to the ground-truth in terms of average displacement.

Adam optimizer~\cite{Kingma2015} with a learning rate of $10^{-3}$ and a batch size of $32$ is used.
After $8$ epochs, the learning rate gets adjusted to $10^{-4}$ and the model is trained for $8$ more epochs.
Similar to LaneGCN, we use a feature size of $128$ for all agent and map encodings.
The classification decoder has a hidden size of $4096$.
For the NMS, we follow an occupancy prediction approach~\cite{Gilles2021} and use $r_\mathrm{NMS} = \SI{1.8}{m}$.
For all models, including the baselines, the optimization during the training is limited to the focal agent only.

All trajectory sets are generated using a subset of $\SI{15000}{}$ randomly picked trajectories from the training dataset.
Instead of generating a separate trajectory set per agent class, we only create one trajectory set for non-vulnerable road users (vehicles, buses, motorcyclists) and one for vulnerable road users (pedestrian and cyclists).

\subsection{Baselines} \label{subsec:baselines}
We use CoverNet and LaneGCN as baselines and reimplement them on Argoverse~$2$.
\textbf{CoverNet} uses a BEV RGB image as the input representation.
This does not allow the consideration of the dynamic state of the vehicle (including e.g., velocity).
\textbf{LaneGCN} encodes the map by only using centerline information.
Agent trajectories are encoded via a series of displacement vectors, meaning that a full tracked state is also not included.
This baseline uses the exact same input information as our proposed RESET method.
In contrast to our set-based method, it directly regresses a fixed number of trajectories.

As described in Section~\ref{subsec:encoder}, to ensure comparability to these baselines, our encoder only makes use of the same input information and does not include the full tracked state and lane information beyond centerlines.

\subsection{Generated Trajectory Sets}
\begin{figure}[!t]
   \vspace{-0.39cm}
	\centering
	\subfloat[Non-vulnerable road users]{%
		\includegraphics[width=0.5\linewidth]{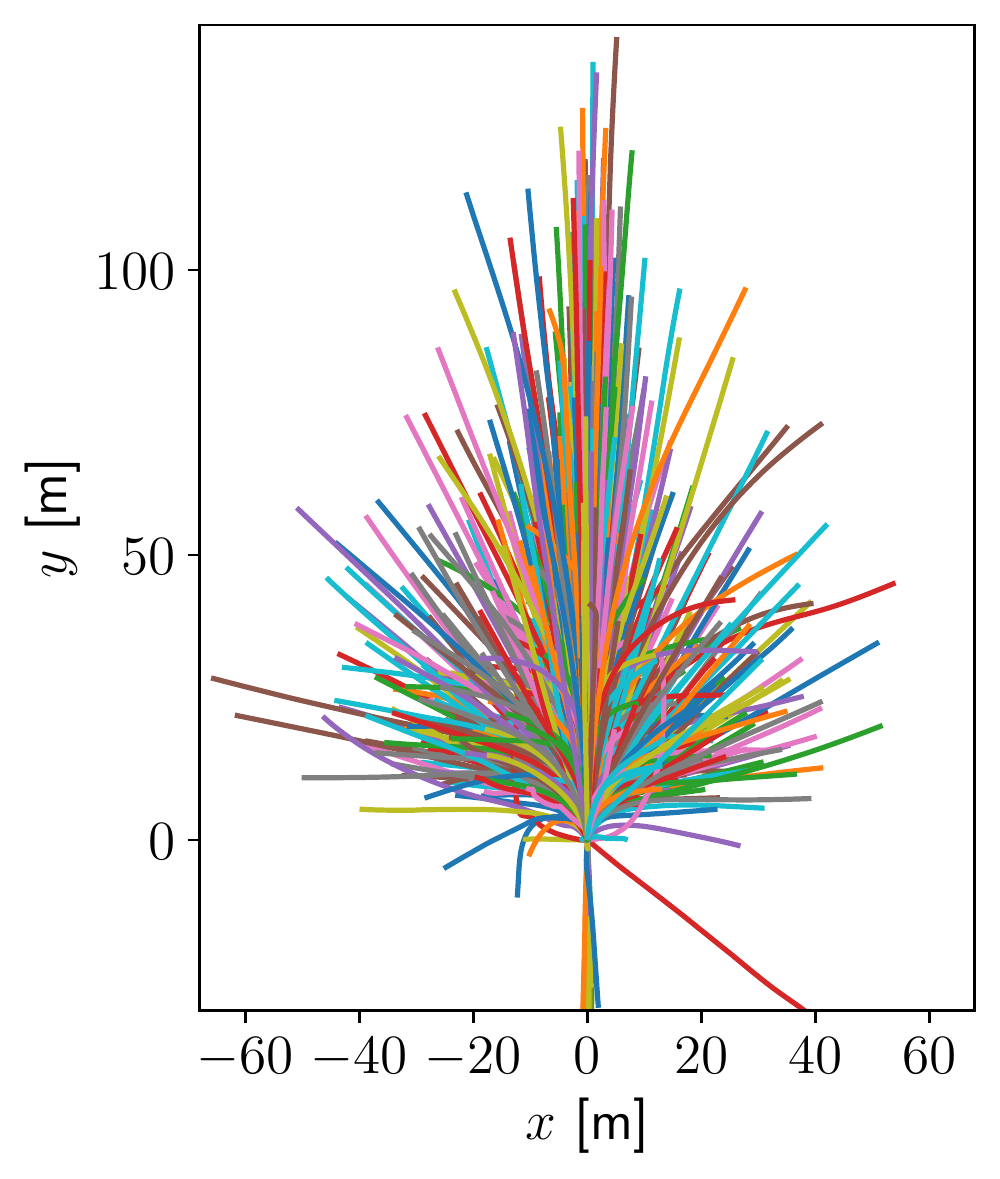}
            \label{non_vulnerable_road_users}}
	\subfloat[Vulnerable road users]{%
		\includegraphics[width=0.5\linewidth]{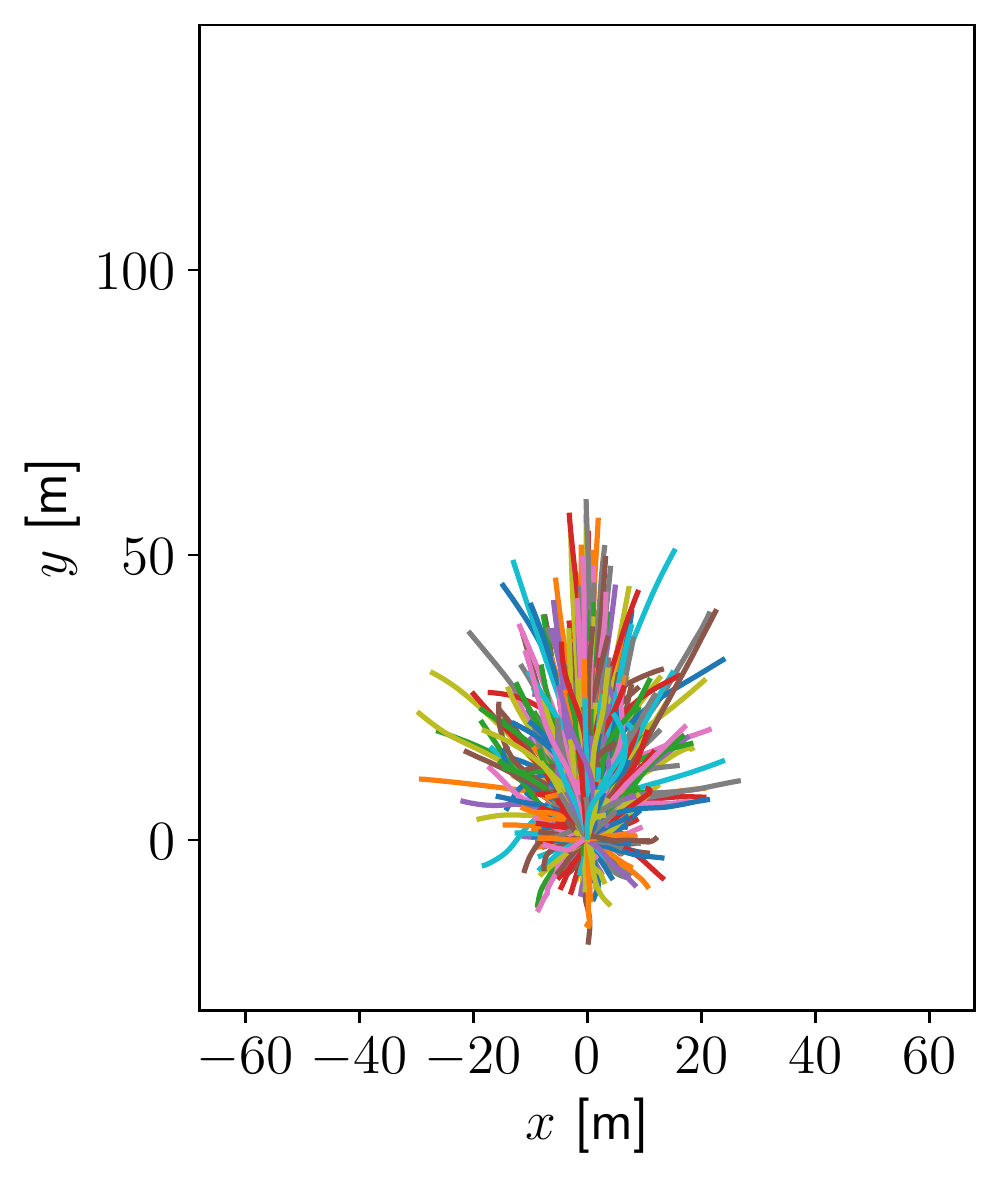}
            \label{vulnerable_road_users}}
	\caption{Visualization of the generated agent class-specific trajectory sets for \protect\subref{non_vulnerable_road_users} non-vulnerable road users and \protect\subref{vulnerable_road_users} vulnerable road users.}
	\label{fig:trajectory_sets}
 \vspace{-0.3cm}
 \end{figure}

Fig.~\ref{fig:trajectory_sets} illustrates the agent class-specific trajectory sets for non-vulnerable road users and vulnerable road users generated with Algorithm~\ref{alg:set_generation}.
The two sets follow different velocity profiles and kinematic models.
Also, there are trajectories aligned into negative $y$ direction that go beyond the plot boundaries.
During the inspection of the Argoverse~$2$ dataset, we found that this was from sequences where the orientation estimate during dataset recording was incorrect.
In Table~\ref{tab:results}, the Lower Bound minADE (LB minADE), which corresponds to the minADE that achievable in case of an optimal classification model, is listed.
A general observation is that our metric-driven way of generating agent-class specific trajectory sets allows for better prediction results, provided that the classification model is able to classify the best trajectories.

\subsection{Evaluation of Prediction Performance}
This section contains quantitative and qualitative results of our method RESET.
Our Python implementation achieves an average prediction time of \SI{17.41}{ms} + \SI{0.48}{ms} (NMS), making it suitable for real-time applications\footnote{Test system specifications: Intel Core i9-7920X, NVIDIA GeForce RTX 2080 Ti.}.
The time without NMS is independent of the number of predicted trajectories.

\subsubsection{Quantitative Results}
\begin{table*}[!t]
	\caption{Quantiative results on the validation split}
	\label{tab:results}
    \vspace{-0.2cm}
	\setlength{\tabcolsep}{2.6pt}
	\centering
	\begin{tabularx}{1\textwidth}{Xlllllllllll}
		\toprule
		\multirow{2}{*}{Model}                       & \multirow{2}{*}{Type} & \multirow{2}{*}{Set size} & \multirow{2}{*}{LB minADE} &                        \multicolumn{4}{c}{$k=1$}                        &                        \multicolumn{4}{c}{$k=6$}                        \\
		                                             &                       &                           &                          & minADE          & minFDE          & MR               & TRI              & minADE          & minFDE          & MR               & TRI              \\ \midrule
		LaneGCN                                      & Reg.                  & -                         & -                        & $\mathbf{2.20}$ & $5.74$          & $\mathbf{67.91}$ & $31.53$          & $\mathbf{0.90}$ & $\mathbf{1.71}$ & $\mathbf{26.30}$ & $27.47$          \\ \midrule
		CoverNet, $\varepsilon = 2$                  & Set                   & $2816$                    & $0.75$                   & $3.46$          & $8.70$          & $78.71$          & $32.09$          & $1.85$          & $3.83$          & $44.09$          & $30.03$          \\
		CoverNet, $\varepsilon = 3$                  & Set                   & $1228$                    & $0.97$                   & $3.24$          & $8.07$          & $79.30$          & $31.22$          & $1.75$          & $3.31$          & $45.65$          & $28.50$          \\
		CoverNet w/ our set                          & Set                   & $1000$                    & $0.74$                   & $3.17$          & $8.09$          & $78.23$          & $26.94$          & $1.74$          & $3.73$          & $46.13$          & $26.74$          \\
		CoverNet w/ our set (agent class-specific)   & Set                   & $1000 {+} 1000$             & $\mathbf{0.73}$                   & $3.08$          & $7.83$          & $77.69$          & $25.88$          & $1.67$          & $3.52$          & $45.32$          & $26.58$          \\ \midrule
		Ours w/ CoverNet set, $\varepsilon = 2$      & Set                   & $2816$                    & $0.75$                   & $2.38$          & $5.84$          & $71.23$          & $29.54$          & $1.42$          & $2.56$          & $33.01$          & $29.01$          \\
		Ours w/ CoverNet set, $\varepsilon = 3$      & Set                   & $1228$                    & $0.97$                   & $2.38$          & $5.71$          & $73.83$          & $29.05$          & $1.52$          & $2.50$          & $37.43$          & $27.75$          \\
		Ours w/ our set                              & Set                   & $1000$                    & $0.74$                   & $2.27$          & $5.59$          & $69.91$          & $25.84$          & $1.34$          & $2.54$          & $35.98$          & $25.92$          \\
		Ours w/ our set (agent class-specific)       & Set                   & $1000 {+} 1000$             & $\mathbf{0.73}$                   & $2.26$          & $\mathbf{5.54}$ & $70.55$          & $\mathbf{25.58}$ & $1.33$          & $2.51$          & $35.90$          & $26.08$          \\
		Ours w/ our set (agent class-specific) + NMS & Set                   & $1000 {+} 1000$             & $\mathbf{0.73}$                   & $2.26$          & $\mathbf{5.54}$ & $70.55$          & $\mathbf{25.58}$ & $1.26$          & $2.28$          & $31.27$          & $\mathbf{24.56}$ \\ \bottomrule
	\end{tabularx}
    \vspace{-0.3cm}
\end{table*}

Table~\ref{tab:results} lists the results of our experiments on the Argoverse~$2$ validation split.
The results of CoverNet show a similar performance for the original CoverNet trajectory set ($\varepsilon=3$) and our metric-driven trajectory set ($1000$).
We argue that the achieved performance is limited by the rasterization-based input representation and the subsequent CNN.
Using agent class-specific trajectory sets ($1000{+}1000$) induces more prior information and therefore slightly improves prediction performance.

Independent of the trajectory set, the graph-based encoder of RESET is able to extract more relevant information about the traffic scene, which is why better results in terms of prediction metrics are achieved throughout.
Also, there is a different trend with regard to the used trajectory sets:
Our metric-driven way to generate trajectory sets outperforms the greedy bagging algorithm used in CoverNet for both, $\varepsilon=3$ and $\varepsilon=2$, by a margin.
Combining it with agent class-specific trajectory sets is superior for single-mode ($k=1$) and multi-modal ($k=6$) prediction.
Furthermore, the subsequent NMS increases prediction performance for multi-modal ($k=6$) prediction even more, leading to a MR reduction by more than $4 \%$.
RESET with agent class-specific trajectory sets and NMS achieves the best values for minFDE$_{@k=1}$, TRI$_{@k=1}$ and TRI$_{@k=6}$.
As is generally the case with GNN-based encoders~\cite{Monninger2023}, our encoder is extendable to make use of more detailed lane information and a full tracked state. Recent approaches making use of such information obtain even lower prediction errors~\cite{Zhang2022_ARXIV}.
Still, our chosen input representation allows for the fairest possible comparison to the baselines.

The direct regression-based approach LaneGCN, making use of the exact same input information, achieves the best multi-modal ($k=6$) prediction results on the Euclidean metrics (minADE, minFDE and MR).
However, it fails to match the set-based approaches in terms of TRI.
One possible explanation is that the output is not restricted to specific feasible trajectories.
Higher values in TRI possibly result in extra burdens for a downstream planner~\cite{Song2021}.

Table~\ref{tab:results_set_size} ablates the performance of RESET for different set sizes.
Using agent class-specific sets with the size $1000$ balances single-mode ($k=1$) and multi-modal ($k=6$) prediction performance while achieving a low TRI.

\begin{table}[!t]
	\caption{Set size ablation study on the validation split}
	\label{tab:results_set_size}
     \vspace{-0.2cm}
	\setlength{\tabcolsep}{1.4pt}
	\centering
	\scriptsize
	\begin{tabularx}{1\linewidth}{Xllllllll}
		\toprule
		\multirow{2}{*}{Set size} &                        \multicolumn{4}{c}{$k=1$}                        &                        \multicolumn{4}{c}{$k=6$}                        \\
		                          & minADE          & minFDE          & MR               & TRI              & minADE          & minFDE          & MR               & TRI              \\ \midrule
		$250{+}250$               & $2.28$          & $5.55$          & $70.69$          & $26.78$          & $1.41$          & $2.39$          & $36.61$          & $25.53$          \\
		$500{+}500$               & $\mathbf{2.23}$ & $\mathbf{5.45}$ & $\mathbf{70.01}$ & $25.91$          & $1.31$          & $\mathbf{2.28}$ & $32.90$          & $24.62$          \\
		$1000{+}1000$             & $2.26$          & $5.54$          & $70.55$          & $\mathbf{25.58}$ & $1.26$          & $\mathbf{2.28}$ & $31.27$          & $\mathbf{24.56}$ \\
		$2000{+}2000$             & $2.29$          & $5.70$          & $70.31$          & $27.17$          & $\mathbf{1.25}$ & $2.32$          & $\mathbf{30.61}$ & $26.15$          \\ \bottomrule
	\end{tabularx}
    \vspace{-0.3cm}
\end{table}

\subsubsection{Qualitative Results}
\begin{figure*}[!t]
	\centering
 	\subfloat{%
		\begin{overpic}[width=0.33\linewidth, keepaspectratio, trim=4cm 2cm 4cm 4.3cm, clip, frame]
			{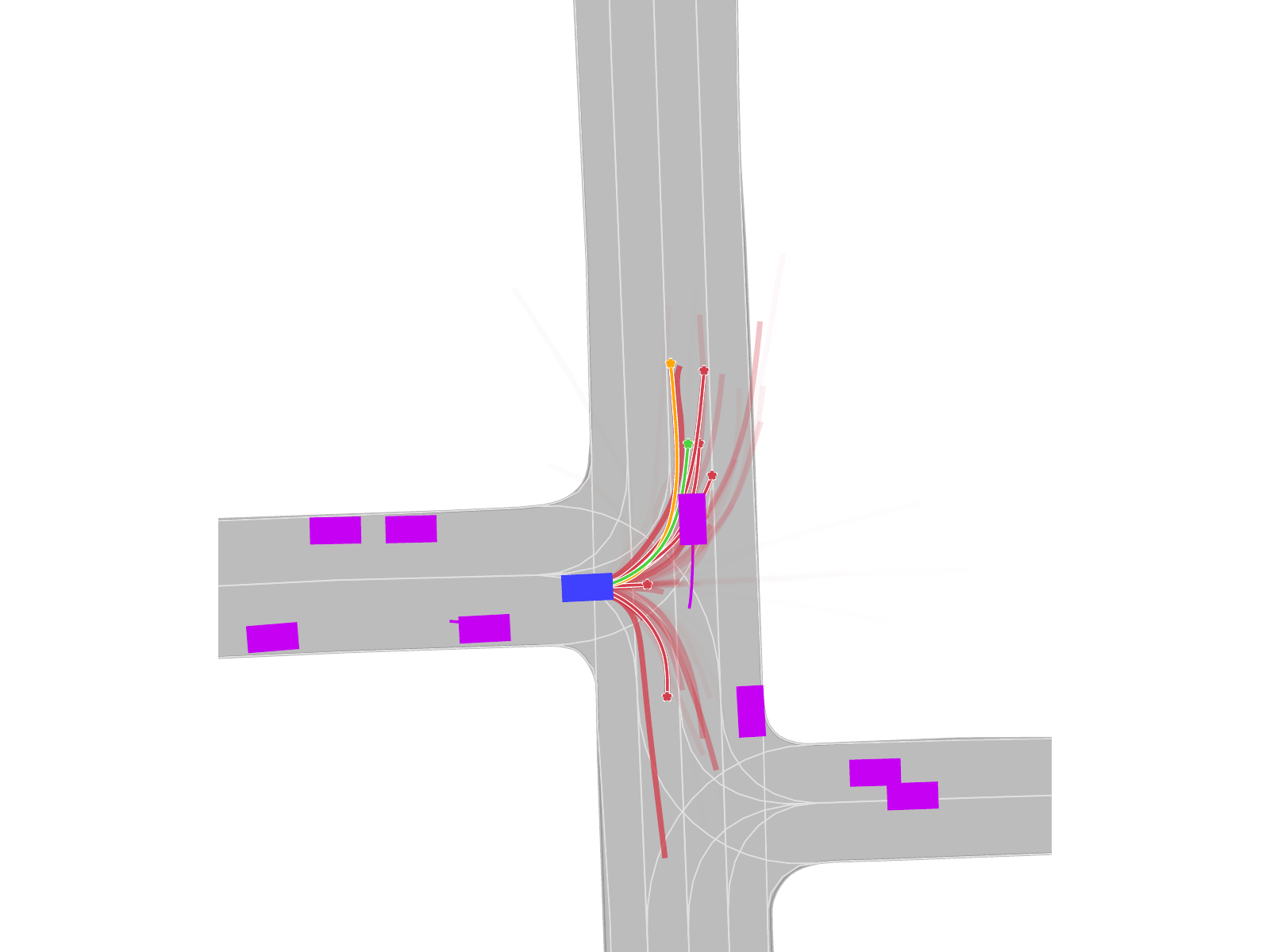}
			\setlength\fboxsep{0pt}
			\footnotesize
			\put(1, 64.6){\pgfsetfillopacity{0.8}\colorbox{white}{\minibox[frame]{\pgfsetfillopacity{1}Seq. $1$}}}
	\end{overpic}}
	\hfill
  	\subfloat{%
		\begin{overpic}[width=0.33\linewidth, keepaspectratio, trim=4.68cm 1cm 4cm 5.79cm, clip, frame]
			{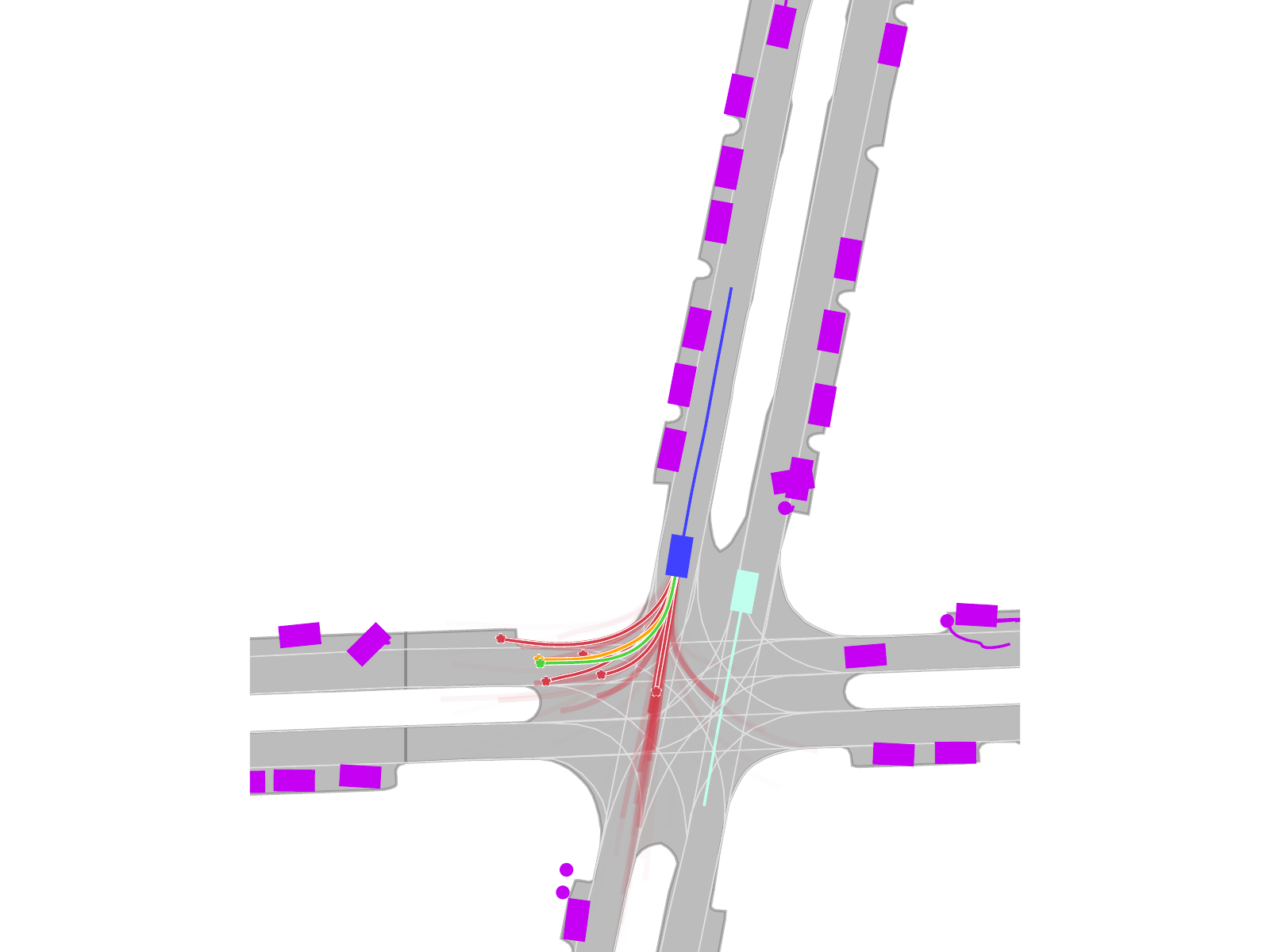}
			\setlength\fboxsep{0pt}
			\footnotesize
			\put(1, 64.6){\pgfsetfillopacity{0.8}\colorbox{white}{\minibox[frame]{\pgfsetfillopacity{1}Seq. $2$}}}
	\end{overpic}}
	\hfill
   \subfloat{%
		\begin{overpic}[width=0.33\linewidth, keepaspectratio, trim=5.6cm 5.4cm 3cm 1.33cm, clip, frame]
			{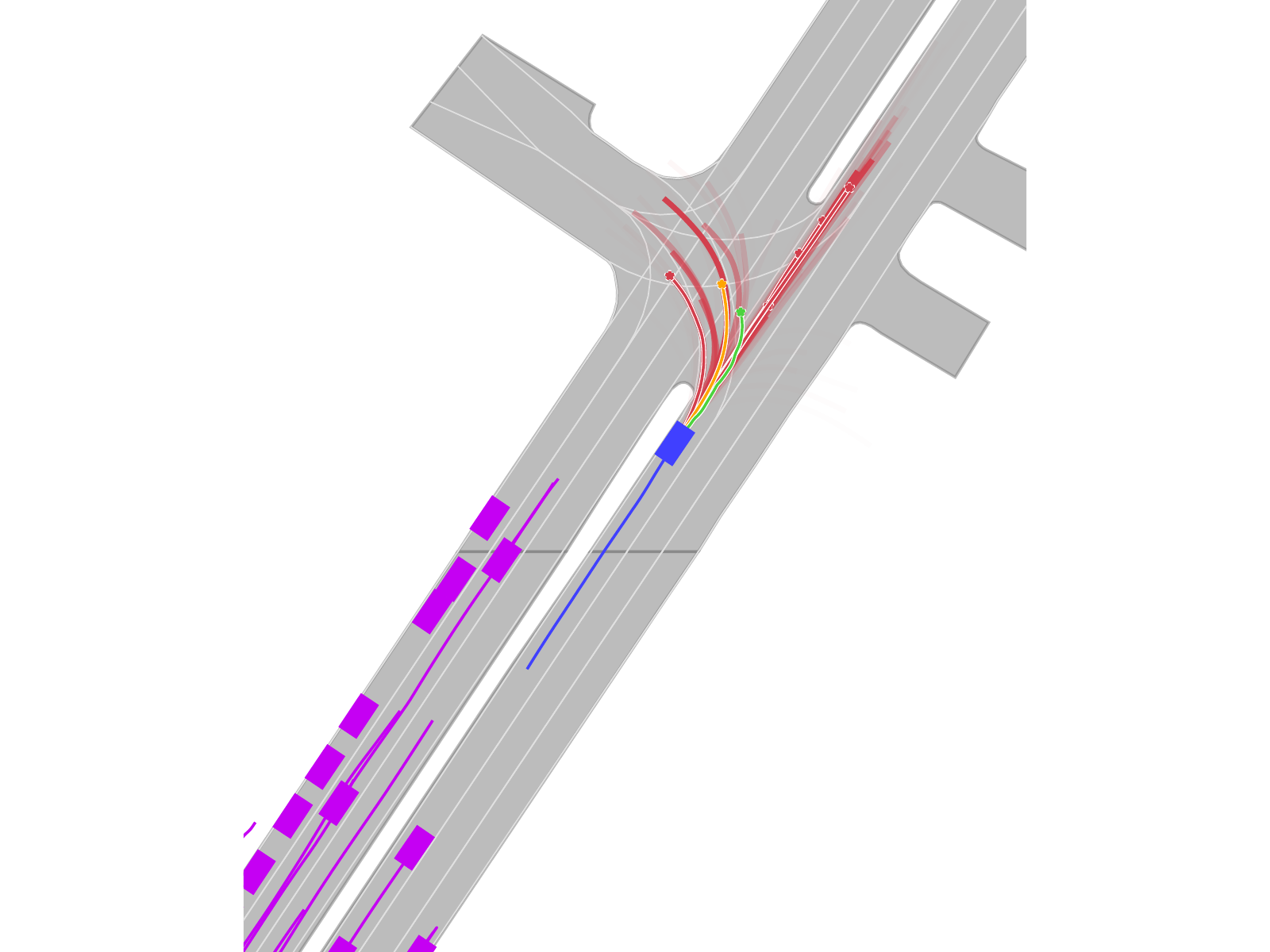}
			\setlength\fboxsep{0pt}
			\footnotesize
			\put(1, 64.6){\pgfsetfillopacity{0.8}\colorbox{white}{\minibox[frame]{\pgfsetfillopacity{1}Seq. $3$}}}
	\end{overpic}}
	\\[-0.29cm]
   \subfloat{%
		\begin{overpic}[width=0.33\linewidth, keepaspectratio, trim=5.6cm 2.7cm 1.0cm 2.7cm, clip, frame]
			{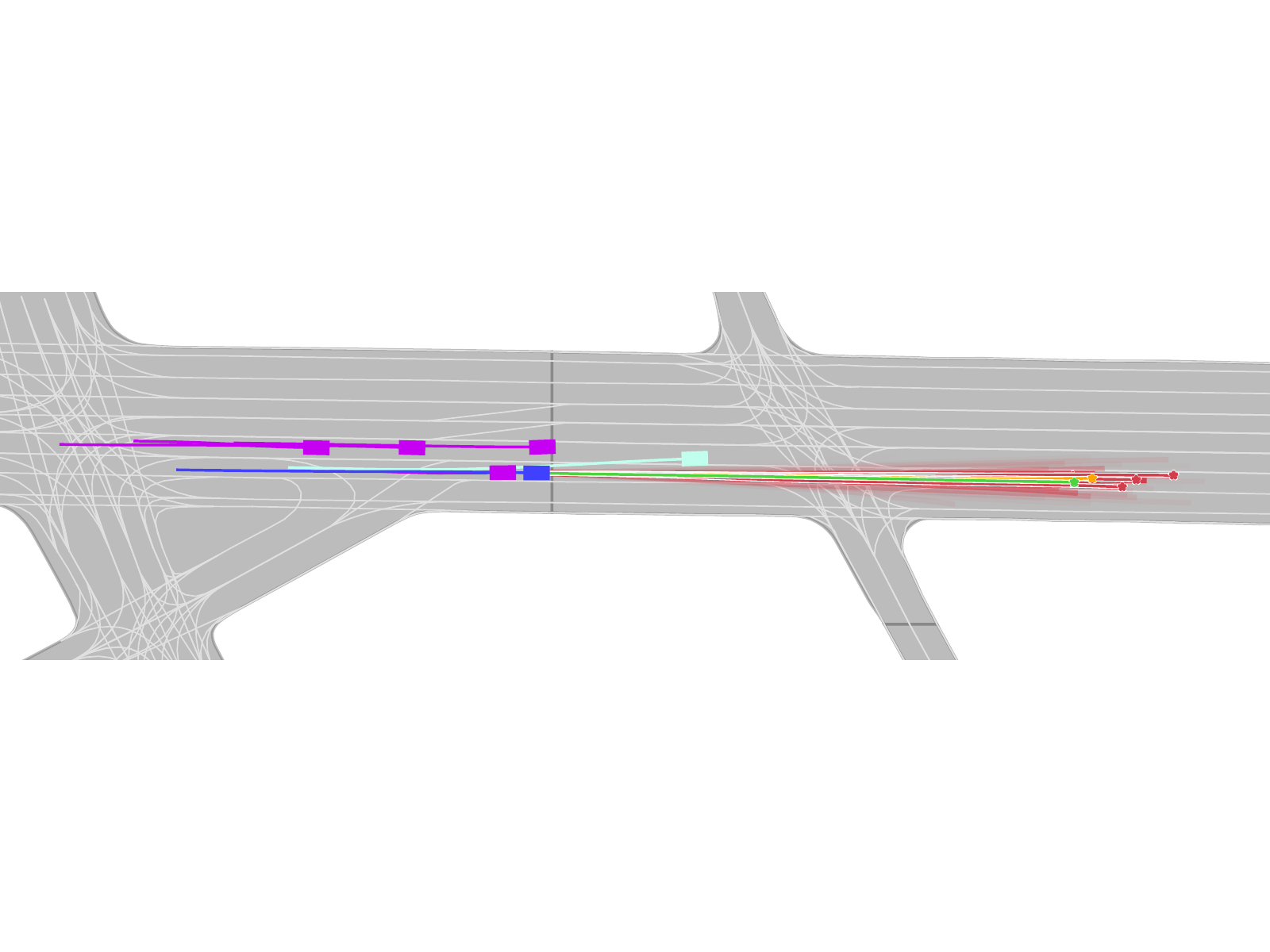}
			\setlength\fboxsep{0pt}
			\footnotesize
			\put(1, 63.6){\pgfsetfillopacity{0.8}\colorbox{white}{\minibox[frame]{\pgfsetfillopacity{1}Seq. $4$}}}
	\end{overpic}}
	\hfill
    \subfloat{%
		\begin{overpic}[width=0.33\linewidth, keepaspectratio, trim=4.8cm 4.3cm 5cm 3.35cm, clip, frame]
			{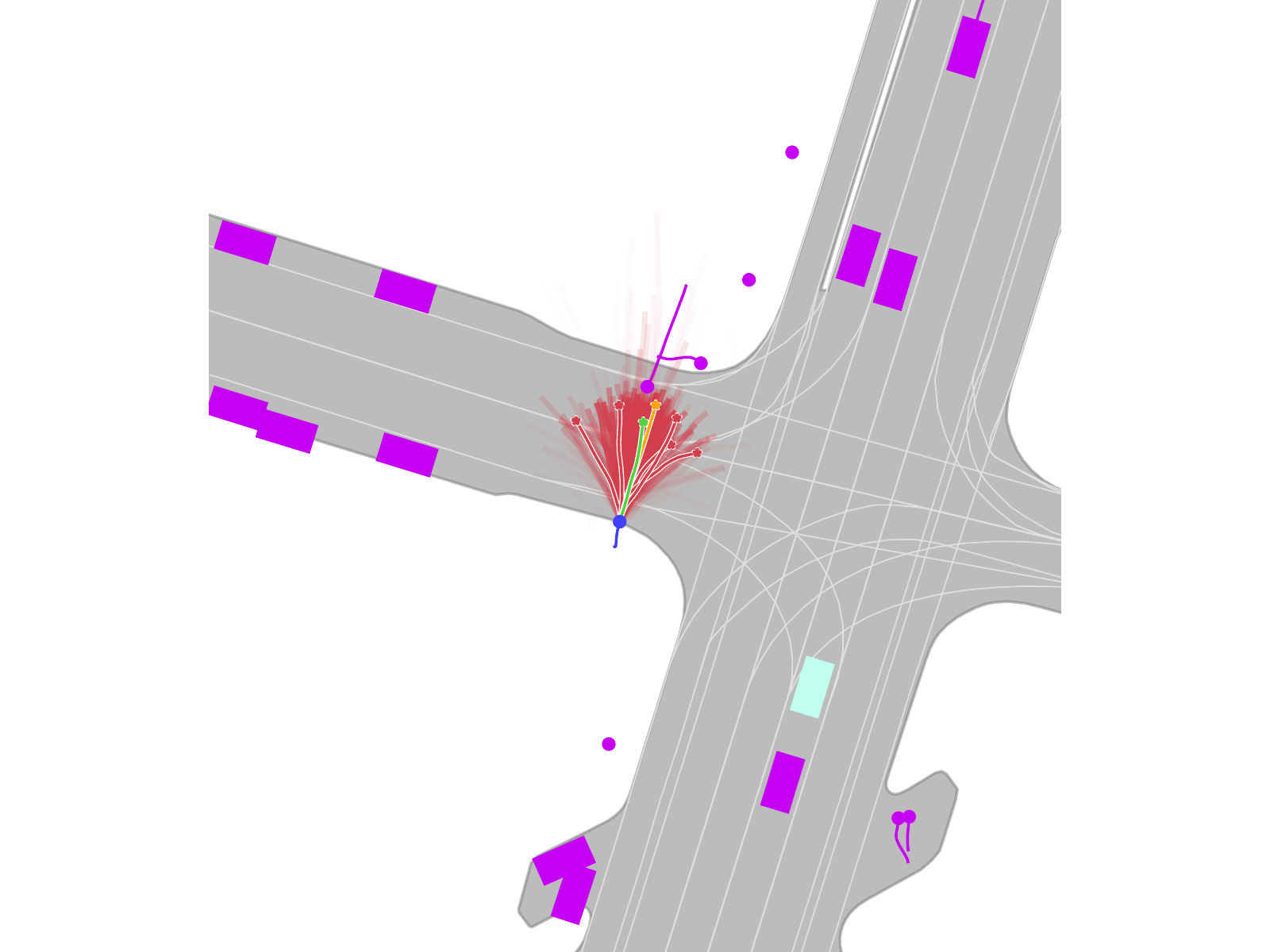}
			\setlength\fboxsep{0pt}
			\footnotesize
			\put(1, 63.6){\pgfsetfillopacity{0.8}\colorbox{white}{\minibox[frame]{\pgfsetfillopacity{1}Seq. $5$}}}
	\end{overpic}}
	\hfill
    \subfloat{%
		\begin{overpic}[width=0.33\linewidth, keepaspectratio, trim=5cm 3.5cm 5cm 4.3cm, clip, frame]
			{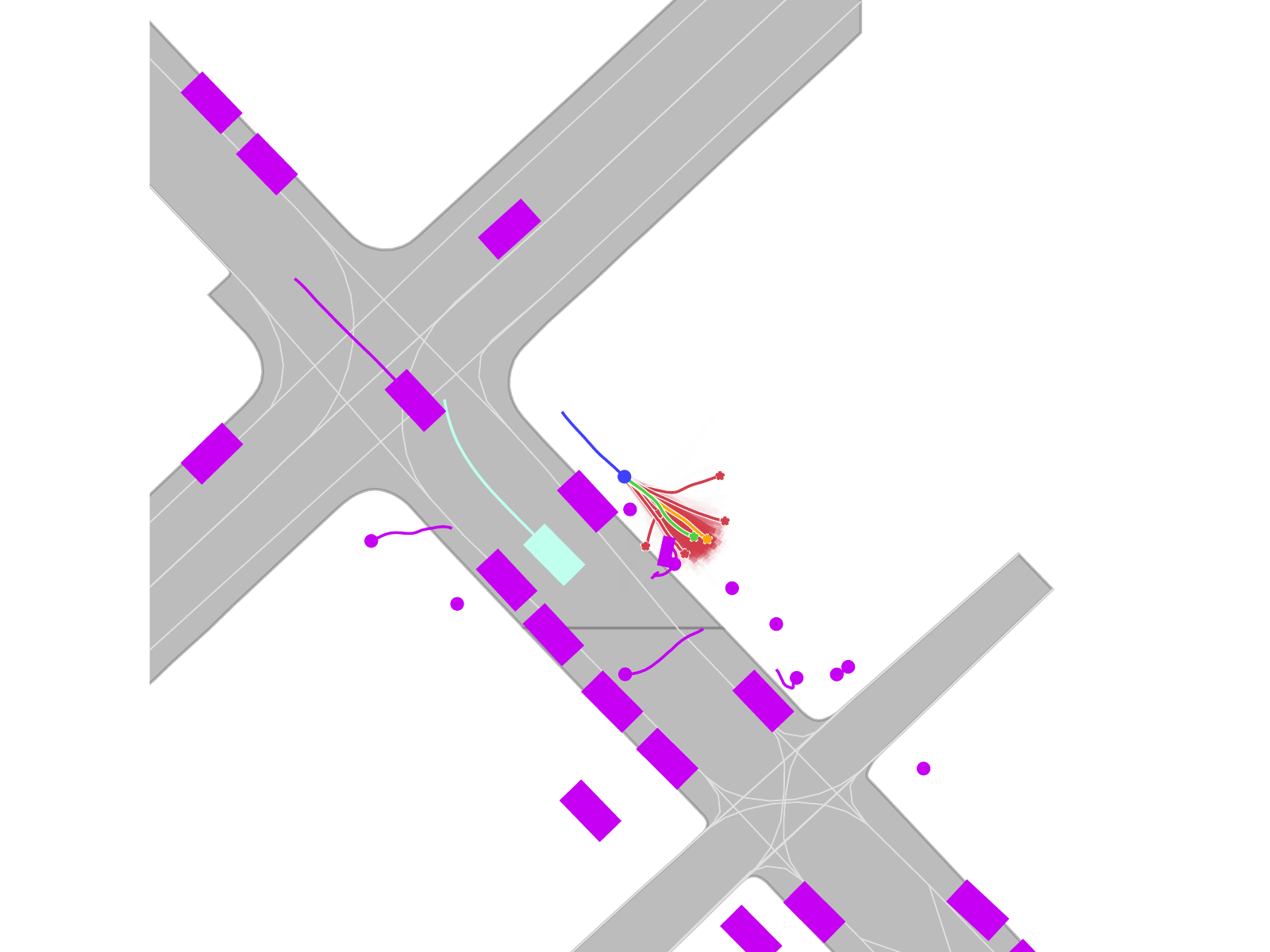}
			\setlength\fboxsep{0pt}
			\footnotesize
			\put(1, 63.6){\pgfsetfillopacity{0.8}\colorbox{white}{\minibox[frame]{\pgfsetfillopacity{1}Seq. $6$}}}
	\end{overpic}}
     \vspace{-0.2cm}
	\caption{Qualitative results of RESET on the Argoverse~$2$ validation set: The past observed trajectory of the focal agent
is colored in blue, the ground-truth future trajectory in green. Predictions are colored in orange and red, with orange corresponding to the most probable
future trajectory. The $k=6$ selected modes are highlighted with a white outline, the remaining trajectories in the trajectory set obtain an alpha value that is proportional to their assigned probabilities. The past trajectory of the autonomous vehicle is colored in cyan and the past trajectories of other vehicles are colored in purple.}
	\label{fig:qualitative_results}
   \vspace{-0.2cm}
\end{figure*}

Fig.~\ref{fig:qualitative_results} shows qualitative results of RESET.
Sequences~$1$ to $3$ illustrate RESET's ability to predict multi-modal trajectories for intersections with different topologies.
Sequence~$4$ is a high-speed sequence, resulting in the predictions being bundled together.
Sequences~$5$ and $6$ show predictions of pedestrians with different velocities, resulting in probability distributions with different characteristics.

\subsection{Evaluation of Conditional Behavior Prediction}
This section contains quantitative and qualitative results of RESET for the task of conditional behavior prediction.
Predictions are conditioned on the ground-truth trajectory of the autonomous vehicle.

\subsubsection{Quantitative Results}
\begin{table}[!t]
	\caption{Conditioning variant ablation study on the validation split}
	\label{tab:results_conditioning}
     \vspace{-0.2cm}
	\setlength{\tabcolsep}{1.5pt}
	\centering
	\scriptsize
	\begin{tabularx}{1\linewidth}{Xlllllllll}
		\toprule
		\multirow{2}{*}{Variant} & \multirow{2}{*}{RCC} &                        \multicolumn{4}{c}{$k=1$}                        &                        \multicolumn{4}{c}{$k=6$}                        \\
		                      &                     & minADE          & minFDE          & MR               & TRI              & minADE          & minFDE          & MR               & TRI              \\ \midrule
		$1$                   & $82.73$             & $4.37$          & $8.88$          & $80.26$          & $\mathbf{24.95}$ & $2.35$          & $3.92$          & $44.18$          & $\mathbf{24.21}$ \\
		$2$                   & $82.73$             & $2.26$          & $5.59$          & $70.28$          & $26.31$          & $1.27$          & $2.31$          & $31.80$          & $24.97$          \\
		$3$                   & $82.69$             & $\mathbf{2.16}$ & $\mathbf{5.28}$ & $68.63$          & $25.89$          & $\mathbf{1.24}$ & $\mathbf{2.23}$ & $\mathbf{30.53}$ & $25.03$          \\
		$4$                   & $64.31$             & $2.17$          & $5.29$          & $\mathbf{68.42}$ & $26.02$          & $1.25$          & $2.26$          & $30.87$          & $25.22$          \\
		$4s$                  & $\mathbf{40.50}$    & $\mathbf{2.16}$ & $5.31$          & $68.67$          & $26.31$          & $1.26$          & $2.26$          & $31.06$          & $25.16$          \\
		$5$                   & $62.38$             & $2.20$          & $5.36$          & $69.00$          & $25.66$          & $1.28$          & $2.28$          & $30.83$          & $24.68$          \\ \bottomrule
	\end{tabularx}
    \vspace{-0.3cm}
\end{table}
Table~\ref{tab:results_conditioning} shows the results of the five variants for conditional behavior prediction.
We introduce a new metric named Remaining Conditional Capacity (RCC).
It corresponds to the ratio between all learnable model parameters used to encode the traffic scene conditioned on the future trajectory of the autonomous vehicle and the total number of model parameters.
A lower RCC is especially advantageous when combining prediction with a downstream planner:
For different planned trajectories of the autonomous vehicle, the values during the forward pass of the prediction module can be reused up to the fusion point.
This means that late fusion is computationally cheaper than early fusion.
Only the percentage of calculation measured by RCC has to be repeated during each forward pass.

Variant~$3$, which uses the AVFuture-To-Agent block prior to the main fusion cycle, achieves the best results in terms of prediction metrics.
Variant~$4$ makes use of a similar AVFuture-To-Agent block, but performs the fusion in later stages of the fusion cycle, which is also reflected by the lower RCC.
However, it still achieves results that are comparable to Variant~$3$.
Prior publications~\cite{Huang2022_ARXIV, Tolstaya2021} that compare early and late fusion for conditional behavior prediction with regression decoders share the same observation:
Early fusion always outperforms late fusion by a margin, with late fusion even leading to worse predictions than without conditioning.
Interestingly, for our set-based method, late fusion obtains similar results than early fusion.
We even test a Variant~$4$s that uses a decoder with a hidden size of $1024$ instead of $4096$ and still achieves comparable performance, while having a significantly lower RCC.
Our set-based method therefore is the first approach with functioning late fusion, making it computationally lightweight to condition the predictions on multiple planned trajectories.

\subsubsection{Qualitative Results}
\begin{figure*}[!t]
	\centering
	\subfloat{%
		\begin{overpic}[width=0.32\linewidth, keepaspectratio, trim=2cm 3.4cm 4cm 1.72cm, clip, frame]
			{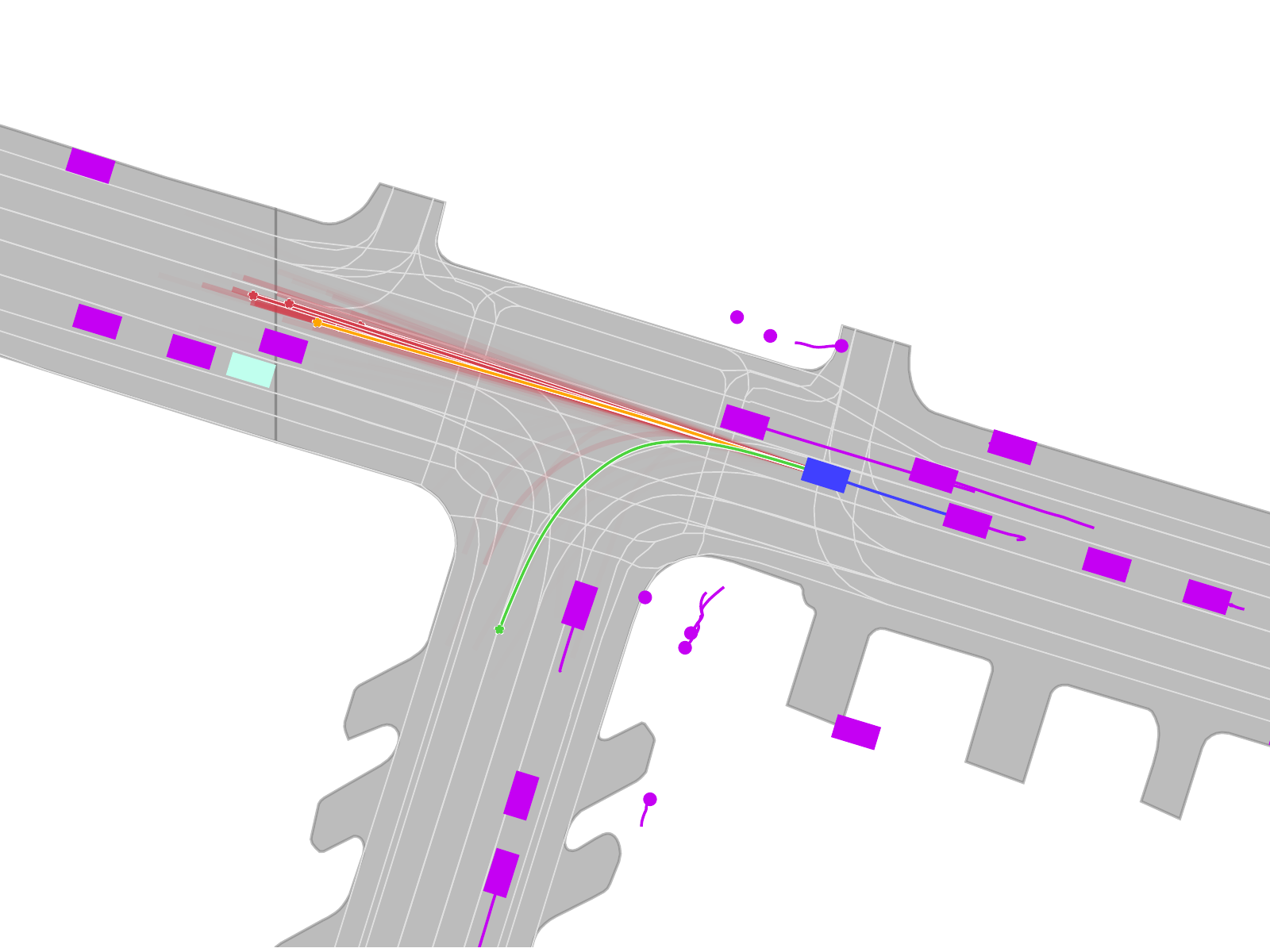}
			\setlength\fboxsep{0pt}
			\footnotesize
			\put(1, 59.3){\pgfsetfillopacity{0.8}\colorbox{white}{\minibox[frame]{\pgfsetfillopacity{1}Seq. $1$\\ Unconditioned}}}
	\end{overpic}}
	\hfill
	\subfloat{%
		\begin{overpic}[width=0.32\linewidth, keepaspectratio, trim=4.3cm 1cm 4.6cm 6.115cm, clip, frame]
			{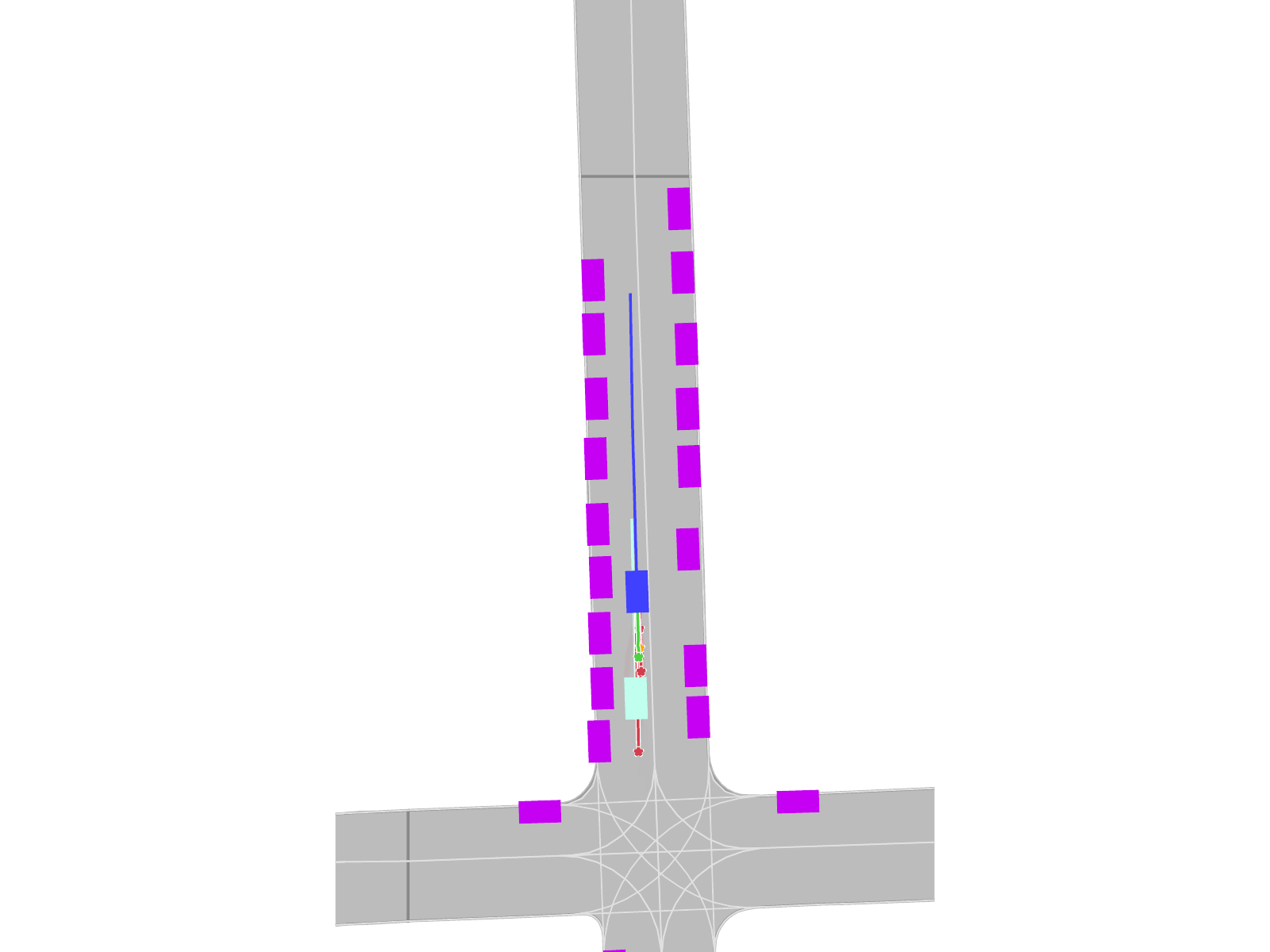}
			\setlength\fboxsep{0pt}
			\footnotesize
			\put(1, 59.3){\pgfsetfillopacity{0.8}\colorbox{white}{\minibox[frame]{\pgfsetfillopacity{1}Seq. $2$\\ Unconditioned}}}
	\end{overpic}}
	\hfill
	\subfloat{%
		\begin{overpic}[width=0.32\linewidth, keepaspectratio, trim=2.62cm 3.2cm 2.62cm 1.4cm, clip, frame]
			{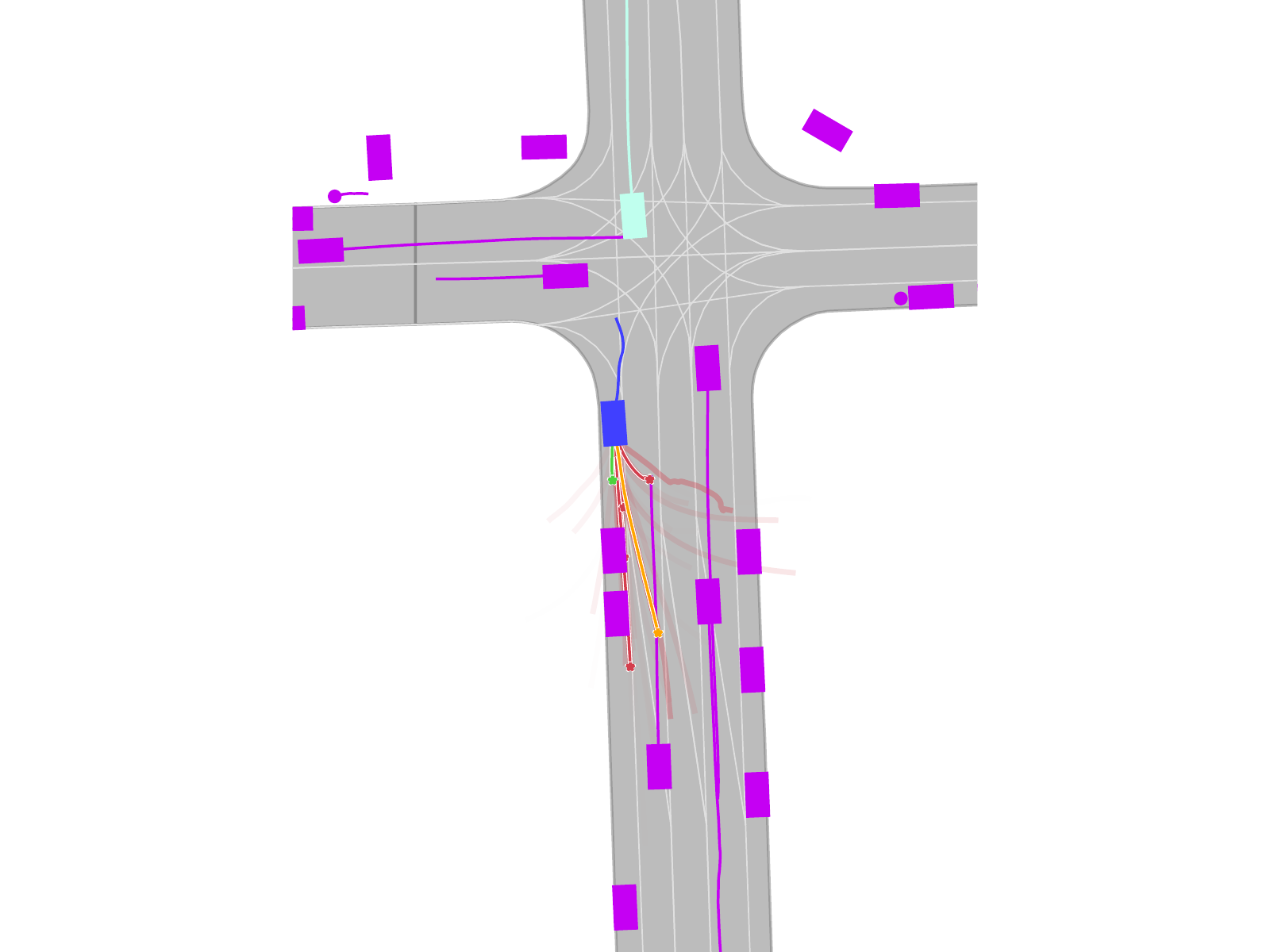}
			\setlength\fboxsep{0pt}
			\footnotesize
			\put(1, 59.3){\pgfsetfillopacity{0.8}\colorbox{white}{\minibox[frame]{\pgfsetfillopacity{1}Seq. $3$\\ Unconditioned}}}
	\end{overpic}}
	\\[-0.29cm]
	\subfloat{%
		\begin{overpic}[width=0.32\linewidth, keepaspectratio, trim=2cm 3.4cm 4cm 1.72cm, clip, frame]
			{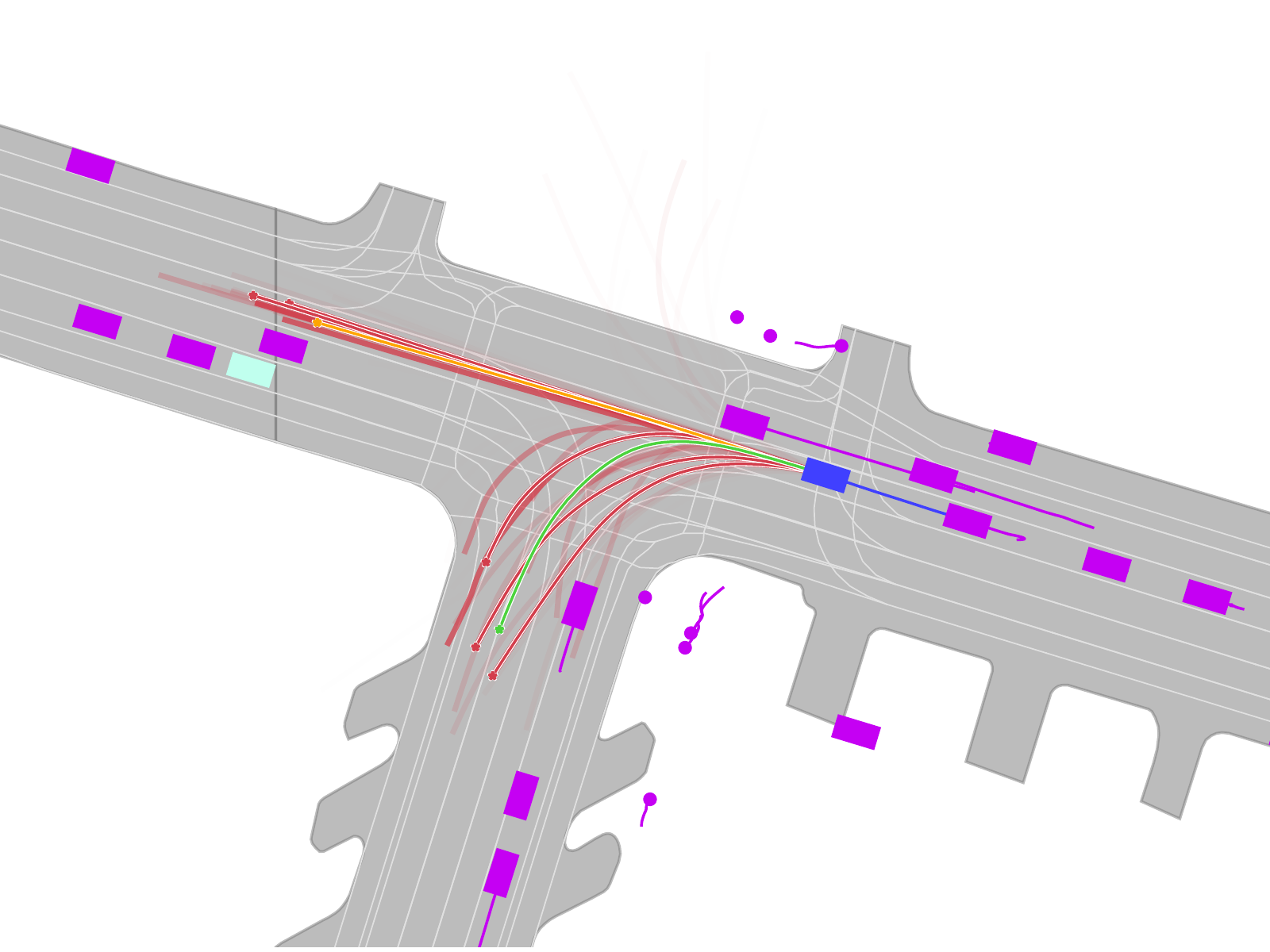}
			\setlength\fboxsep{0pt}
			\footnotesize
			\put(1, 59.3){\pgfsetfillopacity{0.8}\colorbox{white}{\minibox[frame]{\pgfsetfillopacity{1}Seq. $1$\\ Conditioned}}}
	\end{overpic}}
	\hfill
	\subfloat{%
		\begin{overpic}[width=0.32\linewidth, keepaspectratio, trim=4.3cm 1cm 4.6cm 6.115cm, clip, frame]
			{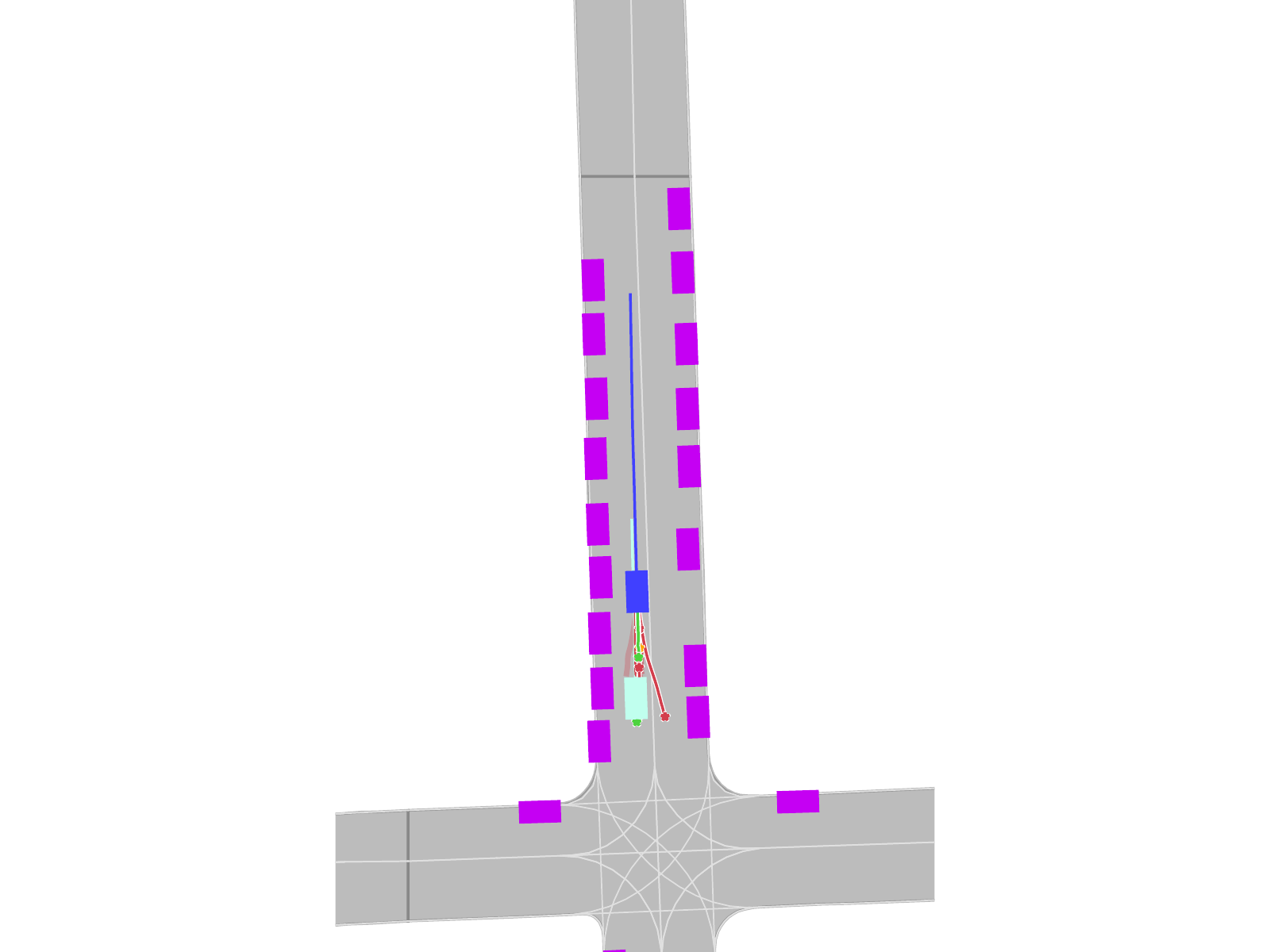}
			\setlength\fboxsep{0pt}
			\footnotesize
			\put(1, 59.3){\pgfsetfillopacity{0.8}\colorbox{white}{\minibox[frame]{\pgfsetfillopacity{1}Seq. $2$\\ Conditioned}}}
	\end{overpic}}
	\hfill
	\subfloat{%
		\begin{overpic}[width=0.32\linewidth, keepaspectratio, trim=2.62cm 3.2cm 2.62cm 1.4cm, clip, frame]
			{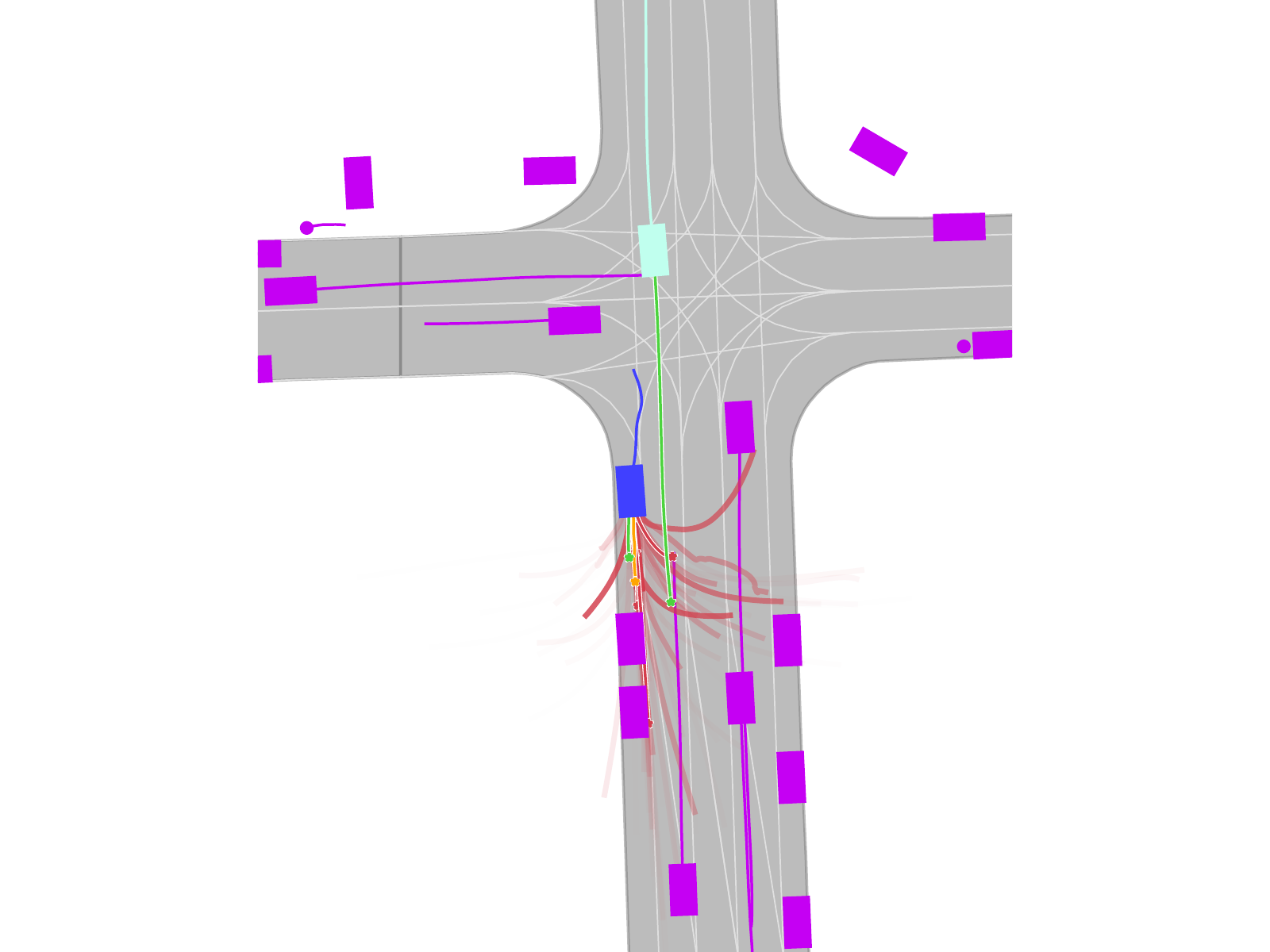}
			\setlength\fboxsep{0pt}
			\footnotesize
			\put(1, 59.3){\pgfsetfillopacity{0.8}\colorbox{white}{\minibox[frame]{\pgfsetfillopacity{1}Seq. $3$\\ Conditioned}}}
	\end{overpic}}
      \vspace{-0.2cm}
	\caption{Qualitative results of RESET on the Argoverse~$2$ validation set without conditional (top) and with conditional behavior prediction (bottom): Color codes are similar to Fig.~\ref{fig:qualitative_results}. The ground-truth future trajectory of the autonomous vehicle (cyan) used for conditioning is also colored in green.}
	\label{fig:qualitative_results_conditioning}
    \vspace{-0.3cm}
\end{figure*}

Fig.~\ref{fig:qualitative_results_conditioning} compares qualitative results of RESET without (top) and with conditioning Variant~$4$ (bottom).
Sequence~$1$ shows a left turn maneuver.
Without conditioning, all predicted modes indicate that the predicted agent will continue driving straight.
With conditioning, the model knows that the autonomous vehicle will yield at the intersection (the autonomous vehicle stands still, hence the future is not visible).
The model concludes that the left turn maneuver is protected, resulting in significantly higher probabilities for left-turning trajectories.
In Sequence~$2$, without conditioning, the agent is predicted to stop behind the autonomous vehicle or crash into it.
With conditioning, the model knows that the autonomous vehicle will barely move forward, resulting in a nudging maneuver being predicted.
In Sequence~$3$, without conditioning, the highest ranked prediction indicates that the agent will change the lane in order to drive around the parked cars in front.
With the additional information that the autonomous vehicle will rapidly approach from behind, the agent is predicted to most likely wait behind the parked cars.
This also corresponds to the ground-truth maneuver.

\section{Conclusion}
This paper employs a set-based method to the task of conditional behavior prediction.
The proposed method RESET combines a new metric-driven algorithm for set generation, a graph-based encoder and an NMS post-processing strategy.
Results for unconditional prediction show that RESET is able to achieve performance comparable to that of a regression-based model.
While late fusion fails in prior work, our set-based method achieves reasonable results with late fusion for conditional behavior prediction.
It remains to combine our conditional set-based method with a downstream planner.
This would allow for a computationally lightweight comparison of predictions resulting from different planned trajectories, allowing the planner to estimate the impact of its decisions on other traffic participants using the prediction model.

\bibliographystyle{IEEEtran}
\bibliography{Literature}

\end{document}